\documentclass[conference]{IEEEtran}
\IEEEoverridecommandlockouts
\usepackage{cite}
\usepackage{amsmath,amssymb,amsfonts}
\usepackage{algorithmic}
\usepackage{graphicx}
\usepackage{textcomp}
\usepackage{xcolor}
\usepackage{hyperref}
\usepackage{multirow}
\def\BibTeX{{\rm B\kern-.05em{\sc i\kern-.025em b}\kern-.08em
    T\kern-.1667em\lower.7ex\hbox{E}\kern-.125emX}}
\begin{document}

\title{Textless NLP - Zero Resource Challenge with Low Resource Compute}

\author {
    \IEEEauthorblockN {
        Krithiga Ramadass$^{\star}$
        \qquad Abrit Pal Singh$^{\star}$ 
        \qquad Srihari J$^{\star}$ 
        \qquad Sheetal Kalyani$^{\dagger}$
    }
    \IEEEauthorblockA {
        $^{\star}$ \textit{Toyota Connected India, Chennai, India} \\
        $^{\dagger}$ \textit{Department of Electrical Engineering, Indian Institute of Technology Madras, Chennai, India}\\
        e-mail: \{krithiga.r, abritpal.singh, srihari.j\}@toyotaconnected.co.in, skalyani@ee.iitm.ac.in
    }
}
\maketitle
\begin{abstract}
This work addresses the persistent challenges of substantial training time and GPU resource requirements even when training lightweight encoder-vocoder models for Textless NLP. We reduce training steps significantly while improving performance by a) leveraging learning rate schedulers for efficient and faster convergence b) optimizing hop length and c) tuning the interpolation scale factors for better audio quality. Additionally, we explore the latent space representation for Indian languages such as Tamil and Bengali for the acoustic unit discovery and voice conversion task. Our approach leverages a quantized encoder architecture, in conjunction with a vocoder which utilizes the proposed mixture of optimized hop length, tuned interpolation scale factors and a cyclic learning rate scheduler. We obtain consistently good results across English, Tamil and Bengali datasets. The proposed method excels in capturing complex linguistic patterns, resulting in clear reconstructed audio during voice conversion with significantly reduced training time.
\end{abstract}

\begin{IEEEkeywords}
unsupervised speech processing, acoustic unit discovery, representation learning, textless NLP
\end{IEEEkeywords}

\section{Introduction}
The availability of text data for low-resource languages has always been a challenge and transfer learning from multilingual models has its own limitations. End-to-End spoken systems without involving text have received significant attention in the recent years. The Zero-Resource challenge (ZRC) \cite{dunbar2022self} has enabled addressing the low-resource language representation problem and has been a significant driver in this area. In the acoustic unit discovery task for ZRC, high-dimensional input speech data is mapped to its latent representation to capture high-level information while discarding the low-level information. \cite{van2021analyzing}. These latent representations are then used as input for downstream tasks, such as the vocoder in a voice conversion task.
Extensive research has been carried out on the self-supervised methods for its improved latent representation. Numerous architectures and its variations are proposed, ranging from small architectures like Contrastive Predictive Coding (CPC) \cite{DBLP:journals/corr/abs-1807-03748} to large transformer based architectures like Wav2Vec \cite{schneider2019wav2vec} and HuBERT \cite{hsu2021hubert}. Similarly, in downstream voice cloning tasks, various architectures like Wavenet \cite{chorowski2019unsupervised}, GANs \cite{kong2020hifi}  and simple RNN/ LSTM architecture \cite{van2020vector} have been proposed and investigated. 
However even the most lightweight architectures require substantial GPU resources. Our focus is on enabling Zero-Resource Challenge namely Textless NLP with low resource compute. We show that we can reduce the training time as well as the required compute resources significantly with our proposed approach while maintaining similar or an  improved performance.
Our major contributions are first, we leverage an appropriately tuned One-Cycle Learning Rate (OCLR) scheduler \cite{DBLP:journals/corr/abs-1803-09820} to reduce the training time—we observe a significant reduction up to 80\% in training steps, thereby accelerating training while maintaining the quality of the reconstructed signal. Then, we improve the performance by further optimizing the hop length and tuning the interpolation scale factors to enhance the audio quality. We also apply this proposed approach on Indian language datasets, specifically Tamil and Bengali. Our experiments demonstrate that having a Vector-Quantized CPC as an encoder followed by an LSTM-based vocoder, incorporating the proposed approach, effectively enables textless NLP tasks with low resource compute across multiple languages.

\section{System Model}
We leverage the Vector-Quantized Contrastive Predictive Coding (VQ-CPC)\cite{van2020vector} as the encoder in our speech processing pipeline. The input audio files are preprocessed and extracted as log-Mel spectrograms. The initial processing involves convolution and normalization layers to extract high-level features. These features are then passed through an auto-regressive network, which predicts future representations of the input based on past information. One of the key characteristics of VQ-CPC is its use of vector quantization as a bottleneck to discretize the continuous embeddings extracted by the autoregressive network into a finite set of discrete codes. This discretization is achieved by mapping each continuous embedding to the nearest entry in a codebook of fixed-size vectors. Compared to more complex architectures like Wav2Vec and HuBert, which utilize transformers in place of the autoregressive component, VQ-CPC offers a simpler architecture. This simplicity contributes to its efficiency and effectiveness, particularly in low-resource language scenarios. For the vocoder, we use a lightweight LSTM-based model. Despite being lightweight in comparison to a Wavenet Autoencoder \cite{chorowski2019unsupervised} or HifiGAN \cite{kong2020hifi}, it still requires significant training time.

\begin{figure}[htbp]
{\includegraphics[width=8cm]{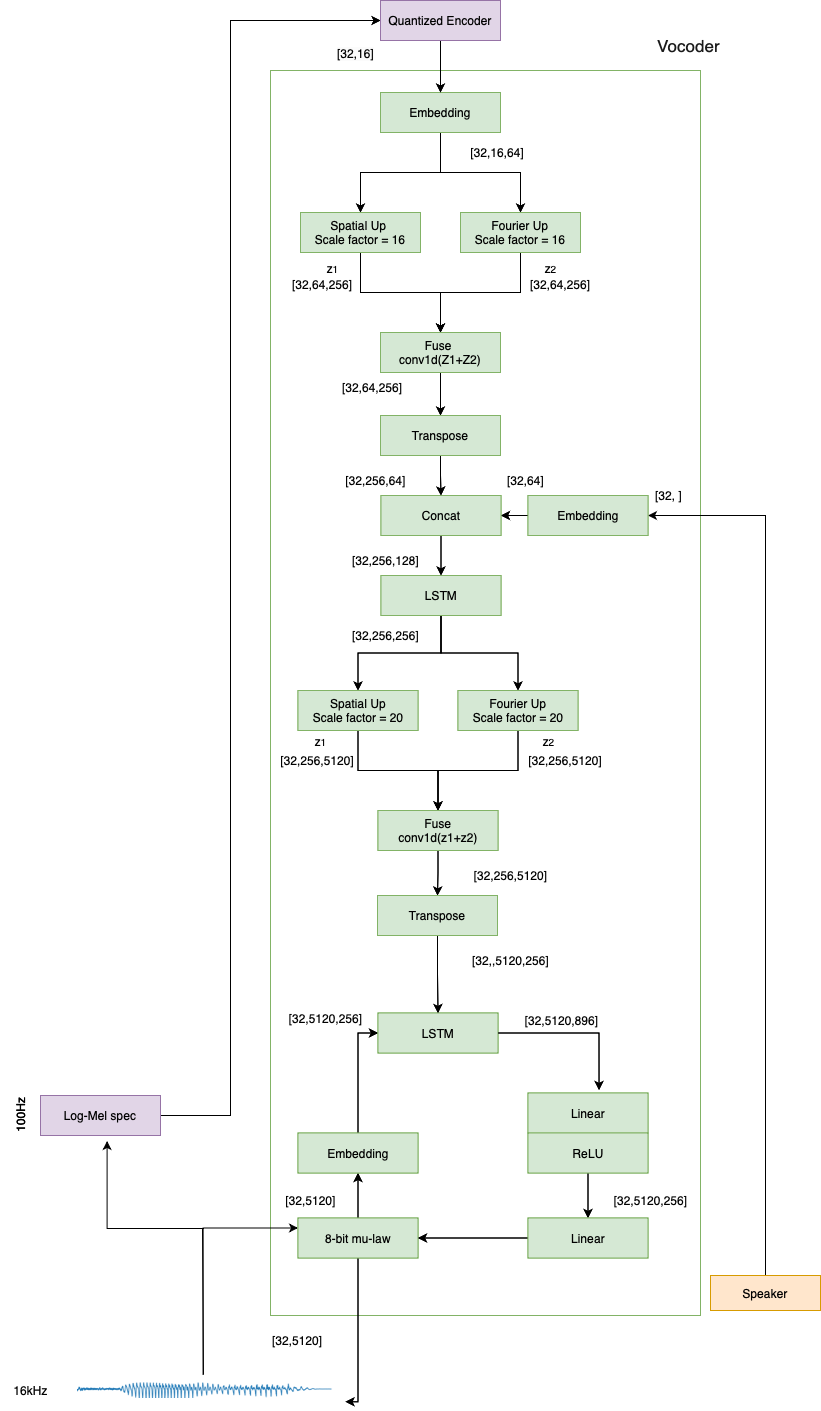}}
\caption{Proposed Vocoder}
\label{fig}
\end{figure}

\begin{table*}[htbp]
\caption{OCLR Results for Encoder and Vocoder (English). Corpus: ZRC 2019}

\label{tab:table1} 
\begin{center}
\renewcommand{\arraystretch}{1.3}
\begin{tabular}{c c c c c c c | c c}
\hline
Model & Steps & LR Scheduler & CER\% & WER\% & PER\% & Bitrate & \multicolumn{2}{c}{ABX}\\
\cline{8-9}
& & & & & & & Within   & Across    \\
\hline
%Baseline 1 & 500K & Multi-step & 39 & - & - & - & 19 & 18\\
Baseline & 160k & Multi-step & 28 & 55 & 46 & 421.39 & 18 & 16\\
Proposed & \textbf{30k} & \textbf{OCLR} & \textbf{27} & \textbf{54} & \textbf{43} & 421.39 & 18 & 16 \\
% Proposed & \textbf{40K} & \textbf{OCLR} & \textbf{26} & \textbf{51} & \textbf{42} & 421.39 & 18 & 16 \\
Proposed & \textbf{60K} & \textbf{OCLR} & \textbf{25} & \textbf{51} & \textbf{40} & 421.39 & 18 & 16\\
\hline
\end{tabular}
\end{center}
\end{table*}

\section{Proposed Approach}\label{PA}
Typically for NLP applications, it requires training for longer duration and hence the Zero-Resource challenge is even more computationally intensive when one is restricted by low compute GPU resources. The vocoder \cite{van2020vector} which is trained for 160k steps on a single T4 GPU instance takes 28 hours of training time. In this work, we try to address this issue of long training time using our proposed combination of learning rate schedulers, smart interpolators, and optimized hop lengths.

\subsection{Learning Rate Scheduler}\label{LRS}
The baseline architecture is a lightweight RNN-vocoder trained with a multi-step learning rate scheduler. \cite{van2020vector}. In this approach, the learning rate was kept constant for the first 50k steps at $4e^{-4}$ and later halved every 25k steps until 160k steps. We observed that loss convergence occurred around 40k steps, but the step decay at constant intervals needed more training steps and that led to inefficient use of resources. Since we wanted to reduce the training time, we resorted to utilize learning rate schedulers which have a cyclic behavior i.e. these schedulers vary the learning rate in a dynamic and non-monotonic fashion. Cyclic learning rate schedulers \cite{smith2017cyclical}, and cosine with warm restarts  \cite{loshchilov2016sgdr} are used extensively in image processing to speed up convergence. 

We used the OCLR approach \cite{DBLP:journals/corr/abs-1803-09820}, where the learning rate starts at a minimum, attains maximum and returns to the minimum during the first few training steps, followed by a decay phase, where the learning rate continues to decay for the rest of the training steps. This phasic training with higher learning rates in the early stages of training allows the model to escape from local minima and explore the parameter space more effectively. Meanwhile, using lower learning rates towards the end of training helped the model converge to a more optimal solution. Determining the base and the maximum learning rate for the cyclic phase is crucial. Using a cyclic learning rate scheduler eliminated the need for an extensive parameter tuning, instead we used a simple Learning Rate Range Test \cite{smith2017cyclical} as a pre-run to the actual training. We performed this test for a very short period starting from a small learning rate value and gradually increasing the learning rate over a few epochs while monitoring the training loss. To understand, how the loss changes as the learning rate is increased at different step rates, we performed three experiments having three step rates 1, 5 and 50. We observed the loss formed a plateau initially, followed by descent and explosion. We identified the peak learning rate for the one-cycle as $4e^{-3}$ from our LRRT experiments. This also helped us identify the optimal learning rate range for our dataset. 

The baseline architecture is RNN-based and suffers from exploding gradient issues, hence we replaced it with a LSTM block. This is required since the RNN-based vocoder is quite sensitive and integrating it with OCLR leads to exploding gradient issues. This lightweight architecture when trained with OCLR scheduler, performed well on English and Indian languages - Tamil and Bengali. We obtained the desired reconstruction quality in just 30k training steps.

\subsection{Tuning Interpolation}\label{TI}
To further improve the quality of the reconstructed output, we experimented with interpolation methods and scale factors in the vocoder. From Fig. 1, one can observe that there are two up-sampling blocks: one before and one after the LSTM block. The first interpolation has an upsampling scale factor of 2 while the second one has an upsampling scale factor of 160. From a signal processing perspective, this is highly imbalanced. To address this, we try to introduce balanced scale factors of 16 and 20, respectively. This simple idea led to a significant improvement in error rates and audio quality metrics. (See Table II for interpolation results for English datasets. See Table III and Table IV for Tamil and Bengali results.) This balanced scale factor contributes to smoother transitions and a more stable speaker's voice.  We also experimented with different interpolation methods. The nearest-neighbor method uses the closest known data point to interpolate, while the linear method estimates values between adjacent data points, assuming a straight line between them. Although we hypothesized that the linear method might yield better results, we did not observe any significant performance improvements.

Further, we recognized that the existing upsampling methods operate solely in the spatial domain. The recent work in \cite{yu2022deep} explores using Discrete Fourier Transform (DFT) to transfer the spatial features to the Fourier domain enabling the spatial-Fourier interaction in the context of general architectures. This method does not replace spatial upsampling but complements it. We implemented tiling and periodic padding in the Fourier domain for our architecture (See Fig 1.) The original implementation \cite{yu2022deep} was designed for images (2D), but since we work with signals, we adapted it using 1D convolutions. This adaptation led to a noticeable improvement in PSNR (Peak Signal-to-Noise Ratio). 
The results are shown in the Table II.

\subsection{Hop length and number of frames}\label{HLNF}

While working on the Fourier upsampling approach, we noticed that using a shorter hop length combined with more sample frames during training yielded better results. A shorter hop length and more sample frames provided more context per batch, improving the training outcomes. We used OCLR to bring down the training time from 28 hours (baseline: 160k steps) to (6,8,12) hours for (30k, 40k, 60k) steps. However, after tuning the hop length, the training time increased to (10,12,19) hours for the same step counts. This increase in training time is relatively less compared to our baseline. Moreover, these changes in the input signal contributed to significantly better evaluation scores (as seen in Table II) and clearer audio output. \newline
In summary, our proposed method leverages signal processing techniques, exploiting them for improved performance. Our method is neither context nor language dependent and we show this by extending our experiments to three very different languages - English, Tamil and Bengali. Note: Tamil and Bengali are low-resource Indian languages in the context of NLP.

\begin{table*}[htbp]
\caption{Vocoder Experiments on Interpolation and Hop Length Experiments (English). LR Scheduler Used: OCLR. Corpus: ZRC 2019}
\label{tab:table2} 
\begin{center}
\renewcommand{\arraystretch}{1.3}
\begin{tabular}{c c c c c c c c | c c c}
\hline
Training Steps & I.Type & Scale Factor & Hop Length & Frames & CER\% & WER\% & PER\% & SSIM & LS-MSE & PSNR \\
\hline
160k (Baseline) & Nearest & 2:160 & 160 & 32 & 28 & 55 & 46 & 0.703 & 0.016 & 17.80\\
\hline
\textbf{30k}& \multirow{3}{*}{Nearest} & \multirow{3}{*}{16:20} & \multirow{3}{*}{160} & \multirow{3}{*}{32} & 27 & 54 & 43 & 0.707 & 0.016 & 17.98 \\
\textbf{40k} &  &  &  &  & 26 & 51 & 42 & 0.707 & 0.017 & 17.75 \\ 
\textbf{60k} &  &  &  &  & 22 & 45 & 38 & 0.714 & 0.015 & 18.16 \\
%\textbf{30K} & Nearest & 16:20 & 160 &  32 & 27 & 54 & 43 & 0.707 & 0.016 & 17.98 \\
%\textbf{40K} & Nearest & 16:20 & 160 &  32 & 26 & 51 & 42 & 0.707 & 0.017 & 17.75 \\
%\textbf{60K} & Nearest & 16:20 & 160 &  32 & 22 & 45 & 38 & 0.714 & 0.015 & 18.16 \\
\hline
\textbf{60k} & Fourier & 16:16 & 128 &  64 & 21 & 44 & \textbf{36} & 0.712 & 0.015 & 18.11 \\
\hline
\textbf{60k} & Nearest & 10:16 &  80 & 102 & \textbf{19} & \textbf{39} & 37 & \textbf{0.719} & \textbf{0.014} & \textbf{18.39} \\
\hline
\end{tabular}
\end{center}
\end{table*}

\begin{table*}[htbp]
\caption{Vocoder Experiments on Interpolation and Hop Length Experiments (Tamil)
LR Scheduler Used: OCLR. Training Steps: 30k Corpus: IISc MILE ASR }
\label{tab:table3} 
\begin{center}
\renewcommand{\arraystretch}{1.1}
\begin{tabular}{c c c c c c | c c c}
\hline
I.Type & Scale Factor & Hop Length & Frames & CER\%  & PER\% & SSIM & LS-MSE & PSNR \\
\hline
Nearest & 2:160 & 160 & 32 & 68 & 76 & 0.638 & 0.03 & 14.78\\
Nearest & 16:20 & 160 & 32 & 66 & 76 & 0.643 & 0.03 & \textbf{14.79}\\
Fourier & 16:16 & 128 & 64 & 65 & 75 & 0.642 & 0.03 & 14.77 \\
Nearest & 10:16 & 80 & 102 & \textbf{61} & \textbf{72} & \textbf{0.653} & 0.03 & 14.69 \\
\hline
\end{tabular}
\end{center}
\end{table*}

\begin{table*}[htbp]
\caption{Vocoder Experiments on Interpolation and Hop Length Experiments (Bengali)
LR Scheduler Used: OCLR. Training Steps: 30k Corpus: Large Bengali ASR }
\label{tab:table4} 
\begin{center}
\renewcommand{\arraystretch}{1.1}
\begin{tabular}{c c c c c c | c c c}
\hline
I.Type & Scale Factor & Hop Length & Frames & CER\%  & PER\% & SSIM & LS-MSE & PSNR \\
\hline
Nearest & 2:160 & 160 & 32 & 34 & 72 & 0.648 & 0.038 & 14.61\\
Fourier & 16:16 & 128 & 64 & 34 & 73 & 0.659 & 0.031 & 15.50\\
Nearest & 10:16 & 80 & 102 & \textbf{31} & \textbf{69} & \textbf{0.667} & \textbf{0.030} & \textbf{15.72} \\
\hline
\end{tabular}
\end{center}
\end{table*}

\begin{table*}[htbp]
\caption{Encoder Results for Tamil and Bengali. Training steps: 22k}
\label{tab:table5} 
\begin{center}
\renewcommand{\arraystretch}{1.1}
\begin{tabular}{c c c c}
\hline
Corpus & BitRate & ABX Within & ABX Across  \\
\hline
IISc MILE Tamil ASR & 442.87 & 29 & 31\\
Large Bengali ASR  & 382.87 & 18 & 28 \\
\hline
\end{tabular}
\end{center}
\end{table*}

\section{Results} \label{Results}

For English, our experiments were conducted on ZRC-2019 and ZRC-2017 datasets. The results shown in Table I are based on ZRC-2019. This dataset includes 102 speakers and consists of 9k train files and 193 test files, amounting to approximately 22 hours of train and test data. The audio files are divided into smaller chunks to enable faster processing. The log-Mel spectrograms are generated and used as input to both the encoder and the vocoder. For Tamil speech corpus - IISc-MILE Tamil ASR \cite{madhavaraj2022subword}, \cite{madhavaraj2022knowledge}, we used a collection of randomly chosen sentences spoken by 112 speakers. We used 37 hours of speech data, with an average duration of speakers varying around 6 seconds per file. In our experiments for Bengali corpus \cite{kjartansson-etal-sltu2018}, we use approximately 40 hours of data collected from 500 speakers. The dataset consists of randomly chosen sentences spoken by various speakers to generalize the linguistic content and speaking styles.

For the voice conversion task, we added two voice samples, duration of an hour which was used as the target speaker's voice for conversion. Similar to \cite{van2020vector}, the speech signals in the three datasets are resampled to 24 kHz. We extract 80-dimensional Mel-spectrogram features using a 40 ms Hanning window, a 12.5 ms frame shift, a 1024-point FFT, and lower and upper frequency cutoffs of 0 Hz and 12 kHz, respectively. The resulting Mel-spectrogram features are then subjected to log dynamic range compression, followed by min-max normalization.

\subsection{Evaluation Metrics}\label{EM}

For encoder, ABX discrimination tests \cite{schatz2013evaluating} and bitrate \cite{chiu2022self} were used to evaluate the discovered acoustic latent units. To evaluate the vocoder, we measured the character error rate (CER) and phoneme error rate (PER). In addition, to measure the audio quality \cite{albadawy2022vocbench}, we used the structural similarity index measure (SSIM), log-Mel spectrogram mean squared error (LS-MSE), and Peak Signal-to-Noise Ratio (PSNR).

\subsection{Results}\label{R}

Table I shows the experimental results for LR scheduler. Our model's switch from the multi-step learning rate scheduler to OCLR (Proposed) gave us significantly better performance. With this change, the number of training steps was drastically reduced from 160k to 30k, or a significant 80\% drop. This resulted in a considerable reduction in the total training duration and a reduction in error rates too. We further extended the training to 60k steps and saw improved performance.

Table II shows the Vocoder experiments for Interpolation and Hop length performed on ZRC 2019 dataset.
We assessed the various interpolation techniques within our learning rate scheduler model aimed at mitigating shrillness, enhancing the speaker's voice stability and improving the overall audio quality. We identified notable improvements in performance associated with interpolation methods. Changing the scale factors from 2:160 to 16:20 improved the performance. Adding Fourier interpolation, further improved the performance. Finally, bringing down the hop length to 80 and setting frames at 102 gave even better outcomes. For the hop length experiment, we adjusted the interpolation scale factor to 10:16 to balance the dimensions of the neural network.

Table III and Table IV shows the vocoder experiments on Interpolation and Hop length for Tamil and Bengali. We used a fine-tuned version of wav2vec2 \cite{baevski2020wav2vec} to calculate WER and PER scores for Tamil and Bengali to obtain transcriptions.\footnote{Baseline scores for Indian languages are not available as the research on these languages is still primitive. Errors from the transcription models might be reflected on the overall evaluation results.} The availability of speech data for these low-resource Indian languages has allowed us to address the lack of sufficient text. In addition, using the proposed tuning combination, the model shows significant results for these languages.
Table V shows the ABX results for Encoder run on Tamil and  Bengali datasets. 
\section{Conclusion}\label{conclusion}
In summary, we show that one can achieve very promising results for textless NLP, even with limited computational time and resource by using a combination of a) smart interpolation, b) tuned hop length and c) appropriate learning rate scheduler. Our approach not only reduces training time but also enhances audio quality, with successful outcomes observed not just for English but also for low-resource Indian languages like Tamil and Bengali. This study also opens the door to re-looking at the various blocks of neural architecture for speech from a signal processing perspective in terms of upsampling, downsampling interpolation  etc., and the possibility of obtaining improved performance with simple changes in these blocks.  Additionally, the method proposed here, to reduce the training time can also be incorporated into more compute-heavy architectures which could be worth investigating in future work.

\clearpage

\bibliographystyle{IEEEbib}
\bibliography{refs}

\end{document}